\pdfoutput=1

\documentclass[11pt]{article}

\usepackage[]{acl}   
\usepackage{times}
\usepackage{latexsym}

\usepackage[T1]{fontenc}

\usepackage{makecell}
\usepackage{amsmath}
\usepackage{graphicx}
\usepackage{epstopdf} 
\usepackage{multirow}
\usepackage{times}
\usepackage{latexsym}
\usepackage{amssymb}
\usepackage{subfigure}
\usepackage{booktabs} 
\usepackage{enumitem}
\usepackage{hyperref} 
\usepackage{tcolorbox}
\definecolor{mybgcolor}{RGB}{227,245,246}  %
\definecolor{myframecolor}{RGB}{119,159,198} %

\usepackage[utf8]{inputenc}

\usepackage{microtype}

\usepackage{inconsolata}

\usepackage{graphicx}

\title{RiTeK: A Dataset for Large Language Models Complex Reasoning over Textual Knowledge Graphs in Medicine}

\author{Jiatan Huang\textsuperscript{\normalfont 1}\thanks{indicates equal contribution}, Mingchen Li\textsuperscript{\normalfont 2}\footnotemark[1], Zonghai Yao\textsuperscript{\normalfont 2}, Dawei Li\textsuperscript{\normalfont 3}, Yuxin Zhang\textsuperscript{\normalfont 2}, \textbf{Zhichao Yang}\textsuperscript{\normalfont 7} \\, \textbf{Yongkang Xiao}\textsuperscript{\normalfont 5}  \textbf{Feiyun Ouyang}\textsuperscript{8}\textbf{,} \textbf{Xiaohan Li}\textsuperscript{6}  \textbf{,} \textbf{Shuo Han}\textsuperscript{8}\textbf{,} \textbf{Hong Yu}\textsuperscript{2,4,8}\\
        \textsuperscript{1}University of Connecticut \textsuperscript{2}University of Massachusetts, Amherst \\
         \textsuperscript{3}School of Computing, and Augmented Intelligence, Arizona State University \\
          \textsuperscript{4}UMass Chan Medical School,\textsuperscript{5}University of Minnesota\\ 
        \textsuperscript{6}Rollins School of Public Health, Emory University \\
        \textsuperscript{7}Optum AI \\
           \textsuperscript{8}University of Massachusetts, Lowell \\
        }

\begin{document}

\maketitle
\begin{abstract}

Answering complex real-world questions in the medical domain often requires accurate retrieval from medical Textual Knowledge Graphs (medical TKGs), as 
the relational path information from  TKGs could enhance the inference ability of Large Language Models (LLMs). However, the main bottlenecks lie in the scarcity of existing medical TKGs, the limited expressiveness of their topological structures, and the lack of comprehensive evaluations of current retrievers for medical TKGs.
To address these challenges, we first develop a Dataset\footnote{The dataset is available here: \url{https://github.com/ToneLi/Medical-Textual-KG-Reasoning-Benchmark/tree/main}} for 
LLMs Complex \textbf{R}eason\textbf{i}ng over
 medical \textbf{Te}xtual \textbf{K}nowledge Graphs (RiTeK), covering a broad range of topological structures. 
Specifically, we synthesize realistic user queries integrating diverse topological structures, relational information, and complex textual descriptions.
We conduct a rigorous medical expert evaluation process to assess and validate the quality of our synthesized queries. RiTeK also serves as a comprehensive benchmark dataset for evaluating the capabilities of retrieval systems built upon LLMs. By assessing 11 representative retrievers on this benchmark, we observe that existing methods struggle to perform well, revealing notable limitations in current LLM-driven retrieval approaches. These findings highlight the pressing need for more effective retrieval systems tailored for semi-structured data in the medical domain.
\end{abstract}

\section{Introduction}

Although large language models (LLMs) have made significant strides in natural language processing (NLP), complex question answering still remains a challenge. 
Medical professionals, for instance, often need to express complex information that combines flexible inputs with specific, structured constraints.
Consider the query, ``\textit{Which organ or tissue function that circulates maternal and fetal blood is affected by Fetal Distress?}’' compared with the simpler version, ``\textit{What does Fetal Distress affect?}’'
Accurately addressing such complex queries is crucial, as it directly impacts healthcare diagnosis and treatment planning.

To effectively answer these queries, organizing the underlying knowledge using medical TKGs becomes essential. TKGs integrate unstructured data, such as textual descriptions of nodes (e.g., the definition of the medical term \textit{Placental Circulation}) with structured data, like the relationships between entities within the graph (e.g., the relationship between \textit{Fetal Distress} and \textit{Placental Circulation} is \textit{affects}).
This integration enables TKGs to represent comprehensive knowledge tailored to specific applications, rendering them invaluable, especially in the medical field, where accuracy and reliability are critically important.

However, existing datasets~\cite{wu2024stark,wu2024avatar} exhibit several critical limitations: they are overly simplistic, typically limited to 1-2 hop reasoning paths; they lack diverse topological structure templates\footnote{The details of the topological structure are provided in the Appendix~\ref{con:Topological_Structures}.} and rich relation types; or they fail to incorporate complex constraints\footnote{Constraints are particularly important in KBQA as they help filter out irrelevant information from large knowledge bases, narrowing the search space and improving both efficiency and accuracy.}.
Consequently, these datasets are inadequate for capturing the complexity of retrieval tasks involving medical TKGs,  where queries demand multi-hop reasoning, diverse topological structure templates, and multiple interdependent constraints.
Moreover, the absence of textual properties in existing medical TKGs limits their effectiveness in delivering comprehensive answers.


To bridge this gap, we introduce RiTek, a large-scale dataset for complex reasoning over medical TKGs. In this progress, one primary technical challenge we address is the accurate simulation of user queries with different reasoning types (e.g., six topological structures in Figure \ref{con:data_flow} within medical TKGs, ensuring that these queries are relevant and reflective of real-world medical scenarios involving patients, doctors, and medical scientists. This challenge stems from the interdependence between textual and relational information, the inherent complexity of medical terminology and relationships, and the limited availability of textual descriptions for medical terms.   We refer to the framework of \citet{wu2024stark} to simulate user queries and construct precise ground-truth answers.
We incorporate richer topological structures that extend beyond the traditional 2- and 3-hop reasoning patterns to better reflect real-world medical scenarios. Compared with datasets like BioKGBench~\cite{lin2024biokgbench} and PrimeKGQA~\cite{yan2024bridging}, the  queries in RiTek not only contains structural information that requires reasoning but also includes textual information related to the ground-truth answer, making the task more challenging. In addition, we enrich the textual descriptions of each node and incorporate more ontology structure.


The key features of RiTeK are summarized as follows: (1) it integrates rich ontological structures and comprehensive textual descriptions, with content quality rigorously validated by medical experts to ensure reliability; (2) it constructs queries that capture complex relational dependencies and nuanced linguistic variations; and (3) it introduces context-sensitive reasoning, where effective retrieval depends not only on a model’s reasoning ability but also on its semantic alignment with entity constraints embedded in the query.

Moreover, we systematically investigate the performance of existing retrieval systems on RiTeK and provide insights to guide future research. In particular, we identify key challenges in processing textual and relational data with complex ontology structures and in mitigating latency issues on large-scale SKBs containing millions of entities and relations.


\section{Related Work}
\textbf{Datasets of Question Answering over Document}. This area of research centers on extracting answers from document sources~\cite{rajpurkar2016squad,dunn2017searchqa,joshi2017triviaqa,trischler2016newsqa,welbl2018constructing,yang2018hotpotqa,jin2021disease,jin2019pubmedqa,hendrycks2020measuring}. For example, SQuAD~\cite{rajpurkar2016squad} assesses a model's ability to interpret and retrieve answers from a single document, focusing on comprehension within a defined context.
PubMedQA~\cite{jin2019pubmedqa} targets reasoning over complex biomedical literature.
MedQA-CS~\cite{yao2024medqa} aims to simulate authentic medical examination scenarios in clinical education.
However, existing unstructured QA datasets often lack the depth required for relational reasoning and failed to address complex user inquiries.
In contrast, our research involves queries that demand more complex relational reasoning, challenging the model’s ability to navigate and utilize structured information effectively.




\noindent\textbf{Datasets of Question Answering over Knowledge Graph}. Structured QA datasets challenge models to retrieve answers from knowledge graphs, which serve as structured databases for factual reasoning \cite{Zhang2017Variational,Yih2016The,gu2021beyond,bao2016constraint,trivedi2017lc}. For instance, MetaQA~\cite{Zhang2017Variational} requires models to infer multi-hop relational paths across entities. To test the models’ abilities to decompose the constraint information in the queries, WebQuestionsSP~\cite{Yih2016The}  is proposed. 
GrailQA~\cite{gu2021beyond} aims to facilitate the answering of more complex questions, as it allows queries to involve up to four relations and optionally includes functions such as counting, superlatives, and comparatives. However, these datasets primarily focus on relational information; the absence of textual context restricts query diversity and limits the semantic expressiveness of reasoning within predefined relationships and entities.


\noindent\textbf{Datasets of Question Answering over Textual Knowledge Graph}.
To integrate textual information into knowledge graphs and queries, the STaRK dataset (Prime, Amazon, Mag)~\cite{wu2024stark} was proposed. To the best of our knowledge, STaRK remains the only dataset that integrates relational and textual information for question answering over TKGs.
However, this dataset exhibits limited topological structure coverage, which restricts its ability to handle complex multi-hop queries, particularly in the medical domain. Furthermore, the lack of detailed node descriptions further impairs a model’s ability to comprehend query semantics. RiTeK addresses these limitations by incorporating richer topological structures and more extensive textual information into both knowledge graphs and queries. This integration leads to more comprehensive and nuanced responses, providing deeper insights drawn from abundant textual data.

\section{Problem Statement}
\paragraph{Textual Knowledge Graph}A Textual Knowledge Graph (TKG) is defined as a graph $\mathcal{G}=(\mathcal{E},\mathcal{R},\mathcal{D})$, where $\mathcal{E}$ denotes a set of entities and $\mathcal{R}$ denotes a set of relations among these entities. In a TKG, the entities and relations are usually organized as {\em facts}, and each fact is defined as a triplet $(h, r, t)$, where $h, t\in \mathcal{E}$ and $r \in \mathcal{R}$ denote the head entity, tail entity and the relation between the two entities, respectively.  Each entity $e$ ($e=h$ or $e=t$) in $\mathcal{G}$ is associated with a textual document $d^e \in \mathcal{D}$ describing its information.

\paragraph{Complex Question Answering over Textual Knowledge Graph}Given a textual knowledge graph $\mathcal{G}$ and input query $q$, the model is expected to generate the answers $a\in\mathcal{E}$, which satisfy the relational constraints defined by the structure of  $\mathcal{G}$  as specified in $q$, and the associated document  $d^e$ needs to satisfy the knowledge required to solve $q$.

 \paragraph{Textual Triple Graph} Unlike traditional knowledge graphs, where each node represents an entity and each edge denotes the relationship between nodes, in the textual triple graph, each node corresponds to a triple (head entity, relation, tail entity) along with the textual description of each entity. In this context, the relation indicates whether the two triples are connected. To be specific, let $\mathcal{G*}=(V,E)$ denote a graph consisting of a set of nodes $V$ and a set of edges $E\in V \times V$. We denote by $n$ the number of nodes in $\mathcal{G}$ and by $m$ its number of edges. Each node $v=(h,r,t,T(h),T(t)) \in V$, where $T(*)$ denotes the textual description of an entity.

\section{Dataset for LLMs Complex Reasoning over 
TKGs (RiTeK)}

\begin{figure*}[t]
        \centering
        \includegraphics[width=1.8\columnwidth]{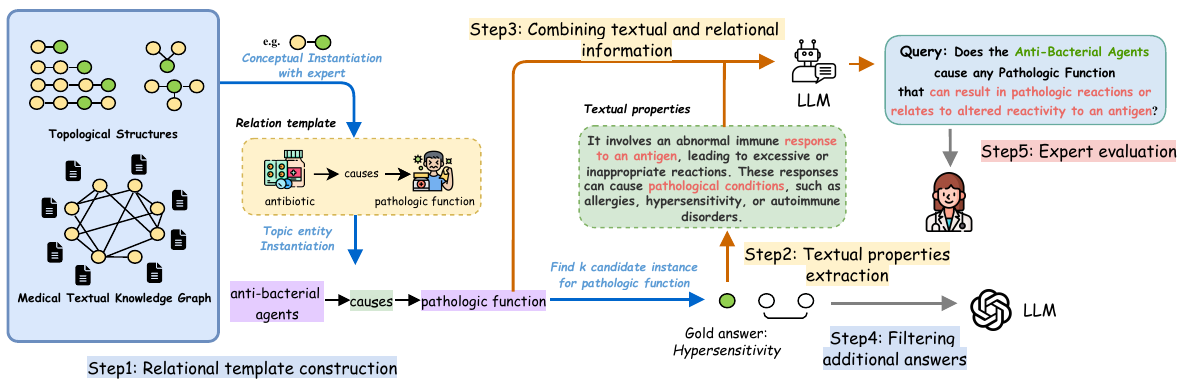}
	\caption{The process of constructing textual structured retrieval datasets involves five main steps, 1) Relational template construction: Create the relation template for TKG using the expert-designed topological structure. 2) Extract Textual Properties:  Choose one node as the answer node that meets the relational requirement, and extract relevant textual properties.
3) Combine Information: Merge the relational information and textual properties to form a natural-sounding query. 4)  Filtering additional answers: Check if the left nodes 
satisfy the textual properties to establish other ground truth nodes. 5) Expert Evaluation: The medical experts evaluate the naturalness, diversity, and practicality of the dataset. }
	\label{con:data_flow}
\end{figure*}

\begin{table}[ht]
	\centering
	\renewcommand\arraystretch{1.3}
	\scalebox{0.65}{
	\begin{tabular} {ccccc}
		\hline 
		
		\hline	
		TKG Dataset& \# Entities  & \# Relation &   \# Triple  &  \# Coverage  \\ 
		\hline		
		STaRK-Amazon&4 &4      & 9,443,802   &--  \\
		STaRK-Mag&4&  4 & 39,802,116  & --  \\
  STaRK-Prime&10 & 18 &  8,100,498 &  15.29\% \\
  \hline
    RiTeK-PharmKG& 3  &   29&  500,958 &   95.61\% \\
    RiTeK-ADint &    102   & 15 &  1,017,284& 36.73\%  \\
		\hline
	\end{tabular}}
		\caption{ Dataset Statistics of constructed medical textual knowledge graphs.  \# Coverage refers to the proportion of nodes with textual descriptions. \# Entities denotes the number of entity types, and \# Relations indicates the number of relation types. As the textual information of the provided nodes is difficult to quantify statistically, we do not include the corresponding statistics for STaRK-Mag and STaRK-Amazon. }
	\label{con:data_statisitics_meidcal_sskg}
\end{table}

\subsection{Medical Textual Knowledge Graph Construction}

We construct two medical TKGs based on PharmKG~\cite{zheng2021pharmkg} and ADint~\cite{xiao2024repurposing}, as the increased number of entity and relation types introduces significant challenges for path retrieval in question answering over textual knowledge graphs.
To enrich entity representations, we incorporate textual details from various databases, including Ensembl, UMLS, and Mondo Disease Ontology. 
As shown in Table~\ref{con:data_statisitics_meidcal_sskg}, our constructed TKGs provide greater node textual coverage, as well as a larger variety of entity and relation types. For further details on these two medical TKGs, please refer to Appendix~\ref{Medical_TKGs}.

\subsection{Question Answering Dataset Construction}
\begin{table}[ht]
	\centering
	\renewcommand\arraystretch{1.3}
	\scalebox{0.65}{
	\begin{tabular} {ccccc}
		\hline 
		
		\hline	
		QA Dataset& \# queries  & \# \makecell[l]{topological \\ structure} &   \makecell[l]{\# instance \\ rate}  & \makecell[l]{train/val/test}   \\ 
		\hline		
		STaRK-Amazon& 9,100 & 1     &   4  & \makecell[l]{0.65/0.17/0.18}   \\
		STaRK-Mag& 13,323&  4 &  1.25      &\makecell[l]{ 0.60/0.20/0.20}\\
  STaRK-Prime& 1,1204 &  3 &  9.3    & \makecell[l]{0.55/0.20/0.25} \\
  \hline
    RiTeK-PharmKG& 1,0235   &6   &  11.33   & \makecell[l]{ 0.80/0.10/0.10 }  \\
       RiTeK-ADint &   5322   &  6& 9.67  &  \makecell[l]{0.80/0.10/0.10  }    \\
		\hline
	\end{tabular}}
		\caption{ Statistical Overview of the Textual KBQA benchmark Datasets. Instance rate refers to the average number of relational templates per topological structure.}
	\label{con:data_statisitics_X_benchmark}
\end{table}

\subsubsection{Overview}
We developed two question-answering datasets,  \textbf{RiTeK-PharmKG} and \textbf{RiTeK-ADint}, based on textual knowledge graphs for complex reasoning. These datasets notably feature queries that integrate relational and textual knowledge, incorporating relational templates with broader coverage and higher instance rates.  Additionally, to enhance their applicability in practical scenarios, these queries mimic real-world query patterns, exhibiting a natural-sounding quality and flexible formats. Specifically,
RiTeK-PharmKG consists of 10,235 synthesized queries. To maximize the coverage of different question topologies, we generate the queries following the six types of topological structure (e.g., multi-hop and constrained multi-hop).  For the synthesized queries, we developed 68 relational templates, crafted by medical experts and detailed in Appendix \ref{relation template in RiTeK-PharmKG}, to encompass various relation types and ensure practical relevance. The instance rate of 11.33, which is higher than that of the current TKG  dataset STaRK (Amazon, Mag, and Prime), highlights the greater diversity of our dataset.
RiTeK-ADint consists of 5322 synthesized queries and covers 6 topological structures, with 58 relational templates. Further details are provided in Appendix \ref{relation_tempalte_RiTeK-ADint}.
To capture the diverse language styles used by different users, we follow STaRK and simulate three distinct roles: medical scientist, doctor, and patient. 
We divide the synthesized queries on each dataset into training, validation, and test subsets,
with the ratios detailed in Table~
\ref{con:data_statisitics_X_benchmark}.
Further details on the scale of our QA benchmarks can be found in Table \ref{con:data_statisitics_X_benchmark}.

\subsubsection{Construction Pipeline}
We present the pipeline used to generates large-scale medical QA datasets on TKGs. The core idea is to
intertwine relational information and textual properties within the queries, accurately constructing ground-truth answers that exhibit more complex topological structures.
The construction of the QA datasets (Figure \ref{con:data_flow}) generally involves five steps, and the specific
processes vary depending on the characteristics of each dataset. These steps are as follows.

\noindent\textbf{Relational Template Construction.}
As shown in Figure \ref{con:data_flow} Step 1, we first created templates based on the 6 designed topological structures~\cite{li2022semantic}, which were evaluated by medical experts to ensure their practical relevance and value.  Afterward, the topological structures are instantiated conceptually with experts.
For instance, for the topological structure \textit{Head entity--relation--tail entity}, the \textit{ "(antibiotic) causes <pathologic function>"} is a valid and common medical relation template, as antibiotics, particularly penicillin and cephalosporins, are well-known for triggering drug hypersensitivity reactions. This makes it a medically reasonable and frequently observed relationship. We then converted these relation templates into specific relationship queries, such as "\textit{Anti-Bacterial Agents} causes \textit{pathologic function}." Since each query could correspond to one or more candidate entities, we matched the queries with the textual KG to obtain k candidate entities.

\noindent\textbf{Extracting Textual Properties.}
 As shown in Figure \ref{con:data_flow} Step 2, for the k candidate answers that meet the relationship criteria, we select one entity as the \textit{gold answer} and use GPT-4 to extract textual properties from the entity's associated document. For instance, in the relationship "\textit{Anti-Bacterial Agents} causes pathologic function," we selected "Hypersensitivity" as the gold answer and extracted its textual properties. These textual properties elaborate on the concept of hypersensitivity, highlighting its key characteristics, which make it more likely to meet the inquirer's needs.

\noindent\textbf{Combining Textual and Relational Information.}
 As shown in Figure \ref{con:data_flow} Step 3, after obtaining the relational templates and textual properties, we combine these components to synthesize the queries. We chose GPT-4 as the LLM for query synthesis, as it excels at generating natural, human-like questions. Additionally, we optimized the prompt and incorporated instructions for different personas to make the queries more diverse and realistic. This approach enhances the quality of our dataset and increases the demands on our model's reasoning capabilities. For details on using GPT-4 to generate this query, please refer to Appendix~\ref{con:prompt of Combining Textual and Relational Information}.

\noindent\textbf{Filtering Additional Answers.} 
As shown in Figure \ref{con:data_flow} Step 4, in addition to the gold answer from which the textual properties are extracted, we need to evaluate whether other candidates meet the requirements of the query in order to include them in the final answer set. We use multiple LLMs to assess whether each candidate's description satisfies the textual requirements of the query. Only candidates that pass validation by all LLMs will be added to the final answer set.

\noindent\textbf{Human Evaluation.}
We invited four medical experts to evaluate 1,000 synthetic queries sampled from two datasets. The evaluation was conducted using a 5-point Likert-like scale across three dimensions. Naturalness measures how grammatically correct and human-like the queries sound. Diversity assesses whether the queries exhibit complex logical structures and encompass multiple entities, relations, and textual requirements. Practicality evaluates the real-world applicability of the generated queries and their likelihood of being encountered in real clinical or everyday scenarios.

The scores were ultimately converted into percentages representing the rates of Positive and Acceptable responses.  We found that the evaluation results provided by GPT-4 for our generated dataset were largely consistent with assessments from medical experts. For shorter queries, such as “What gene is inhibited by naloxone?”, GPT-4 noted the limited relational and textual information contained within and consequently assigned a lower Diversity score. Both GPT-4 and medical experts agreed that certain rare relationship types, such as “an ancestor of”, are infrequently encountered in everyday Q\&A scenarios and are more common in medical education contexts. Only a very small number of queries exhibited issues with insufficient Practicality. The results of this evaluation are summarized in Table \ref{con:data_evaluation_three_dim}. The data in the table represents the Positive/ Acceptable rates (\%) from GPT-4.

\begin{table}[ht]
	\centering
	\renewcommand\arraystretch{1.2} 
 \scalebox{0.8}{
	\begin{tabular} {c|ccc} 
		\hline
		& Naturalness  & Diversity &Practicality  \\ 
		\hline
  	RiTeK-PharmKG &81.80/99.60 & 81.6/99.40 & 67.4/97.8\\
		RiTeK-ADint &81.20/99.20 & 74.80/100 & 68.60/96.60\\  
		\hline
	\end{tabular}}
	\caption{Positive/Acceptable rates(\%) from experts}
	\label{con:data_evaluation_three_dim}
\end{table}

\subsubsection{Data Distribution Analysis}

We chose Shannon Entropy and Type-Token Ratio (TTR) as  metrics to evaluate query diversity generated in our two datasets. Shannon Entropy takes into account the frequency of each word, measuring the evenness of word distribution in the text, while Type-Token Ratio reflects the variety of words, with a higher value indicating greater diversity in the generated queries. 
We found the TTR values for both RiTeK-PharmKG and RiTeK-ADint surpass those of STaRK-Prime, demonstrating that the queries generated in our datasets exhibit high complexity and diversity (The results are shown in Appendix~\ref{con:Data_analysis} and Table~\ref{con:data_statistics_medical_sskg}). For Shannon Entropy, our results are comparable to STaRK-Prime. Since our RiTeK-ADint dataset involves a wide range of non-pharmacological interventions (NPIs), lifestyle modifications, and environmental factors, it introduces a richer variety of specialized terminology and concepts into the synthesized queries. This expanded vocabulary diversity results in significantly higher Shannon Entropy compared to other medical domain datasets. However, since our two datasets are derived from the medical domain, the frequent repetition of specialized medical terminology, as well as the more concentrated vocabulary compared to general-domain texts, results in slightly lower Shannon Entropy for our datasets than for the other two general-domain datasets. For further analysis about the distribution of query lengths  and answer length, please refer Appendix~\ref{Data_Analysis_of_QA_length}.

\begin{table*}[!ht]
	\centering

	\renewcommand\arraystretch{1.3}
\resizebox{1\textwidth}{!}{%
	\begin{tabular} {l|l|ccc|ccc|ccc|ccc|ccc|ccc}
		\toprule 
\multicolumn{1}{c|}  {}&\multicolumn{1}{c|}  {}&\multicolumn{6}{c|}  {\textbf{RiTeK-PharmKG}}& \multicolumn{6}{c|}  { \textbf{RiTeK-ADint}} & \multicolumn{6}{c}  { \textbf{STaRK-Prime}}\\
  \hline
\multicolumn{1}{c|}  {}&\multicolumn{1}{c|}  {}&\multicolumn{3}{c|}  {Exact Match}& \multicolumn{3}{c|}  { Rouge-1}& \multicolumn{3}{c|}  {Exact Match}& \multicolumn{3}{c|}  { Rouge-1} & \multicolumn{3}{c|}  {Exact Match}& \multicolumn{3}{c}  { Rouge-1} \\
\hline
& Approach &P &  R & F1 & P  &   R &F1 &  P &  R & F1 & P  &   R  &F1 & P &  R & F1 & P  &   R  &F1 \\
\hline
    \multirow{9}{*}{Zero-Shot} 
    & GPT-4 & 11.39  &    10.90 & 11.03 & 15.56    & 15.50  &  15.30 &   7.26 &   12.10     &8.03     & 13.71   & 27.64    & 16.35 & 5.23&6.81&4.65&11.31&16.35& 11.31 \\
     &  +Random Walk~\cite{lovasz1993random} &  12.27  &  11.86  & 11.96 &  14.69   &14.15  &  14.30 &   15.12  & 22.68       &   16.52  &20.87    & 32.92    &    23.25   & 7.50&8.20  & 6.48 &13.90 & 17.31& 13.32  \\
       &  +MCTS ~\cite{chaslot2010monte} & 17.17   & 16.54   & 16.68 &19.09     & 18.44  & 18.60 &  16.97   &   24.41     &  18.35   & 22.82   &   34.69  &   25.20      &7.64 & 8.36&6.52 &14.04  &17.45 & 13.38\\
         &  +COT~\cite{wei2022chain} & 13.11  & 16.42  & 13.70 & 17.53  & 22.57  &   18.40 &  10.52  &  19.78      &  11.95   &   17.79 &  37.25   &   20.97   &6.47 &8.23 &5.81 &12.61 &17.99 & 12.47\\
        &  +TOT~\cite{yao2024tree} &   7.31  &    7.32    &7.22    & 13.21   &     14.67    & 13.42   & 3.97   &9.65   & 5.28   & 12.90 & 25.44 & 15.96& 2.99& 3.08&2.55 & 9.50&9.81 & 8.65  \\
   &  +GOT~\cite{besta2024graph} &   3.56  &   4.20    &  3.75  &10.86   & 11.84  & 11.06    & 2.61 &  3.32  &2.81 &  15.09  &  17.63& 15.84   & 1.99& 2.20&1.78 & 9.89& 9.34&  8.72 \\
    &  + TOG~\cite{sun2023think}  & 29.85  & \textbf{38.19}  & \textbf{ 31.14}  & 31.38  & \textbf{40.37}  &  \textbf{32.92}  &  23.08 &\textbf{ 40.63 }    &  25.81  &  27.81  & \textbf{48.93} &31.54  &12.14 &\textbf{ 15.76}& \textbf{11.27}& \textbf{18.67}&\textbf{24.75} & \textbf{18.42 }\\
  



   &  +G-retriever~\cite{he2024g} &  11.21   &    13.39    &   11.60  &  15.01  &   18.54  & 15.62     & 10.97   & 19.05   &12.52 & 17.27   & 32.99  &20.41    & 6.23& 6.61& 5.17&12.01 &14.92 &  11.40 \\
       &  +KAR~\cite{xia2024knowledge}& \textbf{30.95}   & 23.99     &   25.18& \textbf{33.65} &  26.11  &  27.50    & \textbf{39.59 }  &24.00    &\textbf{27.29 }& \textbf{46.54 } & 28.87   & \textbf{32.80}    &\textbf{12.02} &14.49 &11.12 &18.04 & 22.20& 17.61 \\ 
    \hline

    \multirow{9}{*}{Few-Shot} 
    & GPT-4 &  13.75  &  15.54  &14.04  & 16.84    &  19.84 & 17.49  &  17.57   & 17.91       & 17.48    &  25.50  &    28.08 &  26.04 & 7.79 & 6.41& 5.91& 14.03&13.53 & 12.14 \\
    &  +Random Walk~\cite{lovasz1993random} &  11.02  & 13.28   &11.32  &  14.46  & 17.88  &   14.92 &  22.99   &  22.79      &  22.75   & 29.10   &29.07     &    28.95  & 9.93& 6.93& 7.34& 16.54&13.02 &   13.45 \\
      &  +MCTS~\cite{chaslot2010monte} & 17.79  & 17.11   & 17.30  & 20.97    & 20.29  & 20.48 &19.51     &   27.32     &   20.91  &  24.71  &  36.25    &   26.96   &9.57 &6.89 &7.14 &15.92 & 12.55& 12.88   \\
    &  +COT~\cite{wei2022chain} & 17.29   &  16.91 & 16.99 & 21.55    & 20.97  &  21.13 &  18.57   & 18.12       & 18.26    &   26.68 &   26.62  &  26.53   &8.13 & 5.91&5.99 &14.03 & 13.53& 12.14  \\
         &  +TOT~\cite{yao2024tree} &  14.74   &14.74      &14.63   & 19.22  & 19.14   &18.97   &13.28 &  13.17 &13.21 &24.65    &   24.72&  24.60 & 12.84&10.11 &10.36 &6.93 &4.85 &5.06 \\
   &  +GOT~\cite{besta2024graph}  &  12.10  &  12.22      & 12.06   &  17.38  &17.31    & 17.19    & 15.84   & 15.32  &  15.42&26.20   & 25.89  &   25.91  &5.37 & 3.73&3.78 & 12.69&9.98 &10.17  \\
   & +TOG~\cite{sun2023think}& \textbf{ 29.14  } & \textbf{42.33}    & \textbf{ 32.36} & \textbf{ 30.40} & \textbf{44.00}   & \textbf{33.88}     & \textbf{26.50 }  & \textbf{ 47.13  } &\textbf{ 33.83 }&   \textbf{29.46} &  \textbf{49.69 } &   \textbf{36.43 } & \textbf{14.41 } & \textbf{20.39}  & \textbf{16.40}&\textbf{ 19.75 } & \textbf{26.61}&\textbf{ 20.14}  \\



   &+G-retriever~\cite{he2024g} &  12.51   &  12.14     & 12.22    &  15.94  &  15.44   & 15.57    &   17.47&17.50    &17.32  &  24.87   & 24.92  &   24.71  & 7.72& 5.75&5.86  &14.63 &11.92 &  12.10\\
       & +KAR~\cite{xia2024knowledge}  &  27.35 &  27.43    & 26.99  & 29.74  & 29.76    & 29.34    &  34.68  &  33.42  & 33.48&40.15   & 38.55   &  38.88   &13.01 &15.50 &12.21 &19.00 &23.10 & 18.00 \\   
       \hline
  \multirow{3}{*}{Supervised} 
        &G-retriever~\cite{he2024g} & 38.71    & 37.11     &  37.62&  39.78  &  39.18   &  39.31 & 47.93&47.16  & 47.41 & 54.68&  54.00& 54.24  &16.14  &16.47   & 14.11&17.21&27.86     &19.21      \\
       &   GCR~\cite{luo2024graph}&44.38   & \textbf{ 57.28 }  &  47.71 &  46.04  & \textbf{58.83 }   & 49.44     & 43.52  &\textbf{60.78}   &  48.07 &  49.47   &  \textbf{65.57   }& 54.24     & \textbf{19.03}&\textbf{26.89 }& \textbf{18.94}& \textbf{28.01}& \textbf{37.18}&\textbf{ 28.75} \\  
     &   GNN-RAG~\cite{mavromatis2024gnn}& \textbf{ 50.78} & 49.28   & \textbf{49.72 } & \textbf{51.66 }  &  50.29     & \textbf{50.73  }   & \textbf{51.04} & 50.59& \textbf{50.55}& \textbf{56.49}&56.09& 56.09&  16.00 &   15.04 &  14.50 &  24.78   &23.51     &   22.99    \\  
           \bottomrule
	\end{tabular}
 }
 \caption{Results of various approaches for question answering with complex reasoning on RiTeK-PharmKG, RiTeK-ADint and STaRK-Prime. P refers to the Precision, R refers to the recall. In  the experiments, the GPT-4 version is GPT4o-mini.}
\label{con:Model_main_performance}
\end{table*}

\section{Experiments}

\subsection{Retrieval Models and Evaluation Metrics}

We evaluated 11 representative retrieval models on our benchmark datasets under both zero-shot and few-shot learning settings. In addition to our benchmark dataset, we also evaluated the models on STaRK-Prime~\cite{wu2024stark}, a textual question answering dataset with minimal ontological structure in its queries, including:

\begin{itemize}[leftmargin=.1in,topsep=0.1pt]
    \setlength\itemsep{-0.5em}
    \item GPT-4~\cite{achiam2023gpt}: We use GPT-4  with the instruction to generate the answers directly. 
     \item Random Walk~\cite{lovasz1993random}: Starting from the topic entity, a random walk algorithm is applied to explore paths in the textual triple graph in the maximum depth $d$.
       
    \item  MCTS~\cite{chaslot2010monte}: Starting from the topic entity, a Monte Carlo tree search algorithm is applied to explore paths in the textual triple graph in the maximum depth $d$. In this work, we set the $d=3$.
    
    \item Chain-of-Thought (COT)~\cite{wei2022chain}: We designed the instruction to guide GPT-4 in generating the answer step by step, with the output formatted as {step-by-step reasoning: explanation, answer: medical terms}.
     \item Tree-of-Thought (TOT)~\cite{yao2024tree}:We structured the reasoning process as a tree search, where multiple intermediate reasoning paths are explored in parallel. GPT-4 evaluates and expands promising paths based on a voting or scoring mechanism.
     \item Graph-of-Thought (GOT)~\cite{besta2024graph}: We represented the reasoning process as a graph structure, where nodes capture different reasoning states and edges denote transitions. GPT-4 traverses the graph to aggregate information and synthesize the final answer.
  
  \item Think-on-Graph (TOG)~\cite{sun2023think}: is a reasoning framework that enables large language models to interactively perform beam search over knowledge graphs, discovering and evaluating promising reasoning paths without additional training.
   \item G-retriever~\cite{he2024g}: A RAG-based approach that retrieves query-relevant subgraphs using the Prize-Collecting Steiner Tree (PCST) algorithm to enhance LLM understanding and reasoning over textual graphs.
      \item KAR~\cite{xia2024knowledge}: A knowledge-aware query expansion method that augments LLMs with structured document relations from a knowledge graph, using relation-aware filtering to improve retrieval for semi-structured queries.
\end{itemize}

We evaluated the  3 representative retrieval models on our benchmark datasets and STaRK-Prime under supervised learning settings, including:
\begin{itemize}[leftmargin=.1in,topsep=0.1pt]
    \setlength\itemsep{-0.5em}
      \item G-retriever~\cite{he2024g}:A RAG-based approach that retrieves query-relevant subgraphs using the Prize-Collecting Steiner Tree (PCST) algorithm to enhance LLM understanding and reasoning over textual graphs.
    \item  GCR~\cite{luo2024graph}: A knowledge-aware query expansion method that augments LLMs with document-based relational signals to improve retrieval for semi-structured queries.
    \item   GNN-RAG~\cite{mavromatis2024gnn}. A method that uses a GNN to retrieve relevant answers and extract the shortest paths connecting the topic entity and answers, which are then verbalized and fed into the LLM to enhance retrieval-augmented generation (RAG) performance.
\end{itemize}

\noindent We evaluated the outputs of different methods using several metrics, including Exact Match (EM)\cite{rajpurkar2016squad,li2023understand}, which assesses whether the predicted sequence exactly matches the reference, awarding credit only for perfect matches. Additionally, we employed ROUGE-1\cite{cohan2016revisiting} to measure unigram overlap between the predicted and reference sequences, providing partial credit for shared words even when the sequences are not identical. To ensure fairness in the comparison, the instructions and examples are the same for both the zero-shot and few-shot settings, respectively.

\subsection{Results and Discussion}
Table~\ref{con:Model_main_performance} shows the experiment results of  various approaches  based on Exact Match and Rouge-1. We have the following observations. Zero-shot and few-shot setting: (1) We observed that the baseline models struggle to generate the correct answers on RiTeK-PharmKG and RiTeK-ADint. For GPT-4 and GPT+COT, they are challenges in utilizing reasoning information from the graph. Although GPT+COT can utilize step-by-step reasoning, it still relies on the inherent knowledge of the LLM, which limits its ability to apply clear logical reasoning based on knowledge graphs. For the Random Walk, while it can provide reasoning paths, its random nature limits its ability to accurately identify the correct path information. However, it could get the better performance than GPT-4 in RiTek-Adint and STaRK-Prime  in the zero/few-shot setting. (2) Tree-of-Thought (ToT) and Graph-of-Thought (GoT) attempt to guide LLM reasoning through structured prompting, encouraging step-by-step or graph-based logical thinking. However, on complex textual KBQA datasets like RiTeK-PharmKG and RiTeK-ADint, both methods consistently underperform, with F1 scores far below those of retrieval-augmented approaches like KAR (e.g., ToT: 13.42 vs. KAR: 27.50 in zero-shot). This suggests that the internal knowledge and reasoning capabilities of LLMs alone are insufficient for tasks that require fine-grained relational understanding and the integration of attribute information from the query. Despite their logical scaffolding, ToT and GoT struggle to recover factual precision without access to external structured knowledge. (3) KAR achieves strong performance on medical datasets like RiTeK-PharmKG and RiTeK-ADint, outperforming baselines in both zero-shot and few-shot settings. Its main strength lies in combining textual semantics with structured KG relations to generate accurate and context-aware query expansions. However, KAR relies on retrieving the top-n relevant documents; however, determining an appropriate value for n and the optimal order in which to select documents is non-trivial. (4) G-Retriever shows moderate performance across medical datasets, but generally underperforms compared to methods like KAR or TOG in both zero-shot and few-shot settings. For example, on STaRK-Prime, its ROUGE-1 F1 score (5.17 vs. 11.12 zero-shot) lags significantly behind KAR, This indicates a weaker ability to handle complex relational constraints, particularly when the answer’s attributes are embedded in the query. Its main strength lies in interpretable subgraph selection using PCST, which enhances explainability and helps mitigate hallucinations. (5)TOG performs moderately in zero-shot settings but shows strong gains in few-shot scenarios, achieving top-tier ROUGE-1 F1 scores like 37.11 on RiTeK-ADint and 36.43 on STaRK-Prime. This highlights its ability to leverage demonstrations to guide accurate reasoning over knowledge graphs, especially in complex biomedical tasks.

In  the   setting of supervised fine tuning, GCR achieves the best overall performance across all three medical benchmarks in the supervised setting, with scores like 57.28 ROUGE-1 F1 on ADint and 49.72 on STaRK-Prime, demonstrating its strength in generating faithful, KG-grounded answers. However, GCR relies on pre-constructed KG-Trie indices. We found that GNN-RAG achieves better performance on the RiTeK-PharmKG and RiTeK-ADint datasets, demonstrating its ability to retrieve relevant path information from the graph. However, since it primarily relies on shortest paths, it may overlook critical reasoning information embedded in more complex or indirect graph structures.

\begin{table*}[!ht]
	\centering

	\renewcommand\arraystretch{1.3}
\resizebox{1\textwidth}{!}{%
	\begin{tabular} {l|l|ccc|ccc|ccc|ccc|ccc|ccc}
		\toprule 
\multicolumn{1}{c|}  {}&\multicolumn{1}{c|}  {}&\multicolumn{6}{c|}  {\textbf{llama 3.1 8b}}& \multicolumn{6}{c|}  {\textbf{llama2-chat-7b}}& \multicolumn{6}{c}  {\textbf{Biomixtral 7b}} \\
  \hline
\multicolumn{1}{c|}  {}&\multicolumn{1}{c|}  {}&\multicolumn{3}{c|}  {Exact Match}& \multicolumn{3}{c|}  { Rouge-1}& \multicolumn{3}{c|}  {Exact Match}& \multicolumn{3}{c|}  { Rouge-1} & \multicolumn{3}{c|}  {Exact Match}& \multicolumn{3}{c}  { Rouge-1} \\
\hline
    &Approach &P &  R & F1 & P  &   R &F1 &  P &  R & F1 & P  &   R  &F1&  P &  R & F1 & P  &   R  &F1\\
\hline
    \multirow{3}{*}{ RiTeK-PharmKG} &  G-retriver~\cite{he2024g}   &  \textbf{ 36.97}   & \textbf{46.07} & \textbf{38.31}  & 33.04  &  47.03  & 38.41  &   38.71  &   37.11  & 37.62  &  39.78    & 39.18    &  39.31   &  \textbf{43.01 } & \textbf{41.59}   &  \textbf{   42.01  } &  \textbf{  43.95 }  & \textbf{42.69} & \textbf{ 43.10}   \\
    &  GNN-RAG & 33.21  &  43.01    &  37.31&   21.00  &  44.89 &  26.00    & \textbf{ 50.78}  &   \textbf{ 49.28}   &   \textbf{49.72 }  &   \textbf{51.66 }  &   50.29   &  \textbf{50.73} & 39.93  &   39.12   &    39.26 &    41.69 &   40.89  & 41.08\\
    
& w/o  retriever  & 32.45  &  43.40 & 34.23  &  \textbf{47.60 } &   \textbf{46.59}   &  \textbf{46.84}   &     38.91   &  37.63   &  38.02   &  40.57   &  39.31  & 39.72  &  41.49    &  39.43   &  40.05   &  41.25   & 41.06&40.99\\
         \hline
          \multirow{3}{*}{ RiTeK-ADint}  &  G-retriver~\cite{he2024g}   &  \textbf{50.83 }    &  \textbf{50.07 }&   \textbf{50.31}   & \textbf{  57.34} &  \textbf{56.61}   &  \textbf{56.87}  &          47.93   &   47.16  &  47.41 &  54.68    &   54.00  &   54.24  &  48.34 &   47.48   & 47.75    &   55.58  & 54.73 & 55.02   \\
    &  GNN-RAG & 40.88  &  40.90    &40.43  &  44.43   &45.01   & 45.49        & \textbf{ 51.04  }&   \textbf{50.59}    &   \textbf{50.55  }  &   \textbf{56.49}   &  \textbf{ 56.09 }  & \textbf{ 56.09} & \textbf{ 50.83  }&   \textbf{ 50.07 }  &   \textbf{50.31 }  & \textbf{  57.34}   &  \textbf{ 56.61 }   &  \textbf{56.87}\\
    
  
& w/o  retriever   & 49.59  & 48.48  & 48.82  & 55.23  &  54.29   &  54.61  &        46.58   &   45.82  &  46.06   &   51.66  &  49.91  & 46.47  &   49.79   &  48.93   &  49.20   &  56.43   &53.80&54.15 \\
\hline
   \multirow{3}{*}{ STaRK-Prime}  & G-retriver~\cite{he2024g}   &  \textbf{ 16.14 }   & \textbf{ 16.47} &  \textbf{14.11   } & 17.21  & 27.86   & 19.21  &     10.15 &    8.45 &  8.17 &  21.75    &  18.08   &  18.40   & 12.22  &   \textbf{ 11.54 }  &  10.58   &   23.15  &    21.37&20.72\\
     & GNN-RAG & 7.81  &   16.67   & 9.35 &  \textbf{ 18.13  } &   \textbf{27.50} &  \textbf{ 19.65}    &   \textbf{16.00} &   \textbf{15.04   }  &   \textbf{ 14.50}  &  \textbf{ 24.78}   &  \textbf{ 23.51}   &  \textbf{22.99} & 11.20  &     10.31 &  10.65   &  17.98   & 18.09    & 18.32\\ 
  
& w/o  retriever &  12.96 &  14.99  & 11.91  & 16.80  &  25.75   & 18.41  &    11.77  &  10.38   &   9.68  &  20.65   & 21.59   &  17.66 &   \textbf{12.96}    &  11.12   &   \textbf{10.73 }  &  \textbf{ 24.83 }  & \textbf{21.59}& \textbf{21.68} \\

           \bottomrule
	\end{tabular}
 }
 \caption{Performance of different retrieval models across backbone LLMs. “w/o retriever” denotes an LLM  without retrieval augmentation.}
\label{con:Model_different_llm}
\end{table*}  %
\subsection{Analysis}
\subsubsection{Effect of Different LLMs on Retriever Effectiveness}
In this part, we analyze the influence of different LLMs on the retrievers.
Table~\ref{con:Model_different_llm} presents the performance of three retrieval settings, G-retriever, GNN-RAG, and without retriever, in three LLMs of the backbone: Llama 3.1 8b, Llama2-chat-7b, and Biomixtral 7b, on three datasets. Overall, G-retriever consistently outperforms other approaches across most metrics, particularly in Rouge-1 F1 scores. For instance, on RiTeK-ADint, G-retriever with Llama 3.1 8b achieves the highest F1 score of 56.87, while the GNN-RAG and no retriever baselines lag behind. Similarly, G-retriever reaches 55.02 F1 on Biomixtral for the same dataset, showcasing its robustness across model sizes. In contrast, GNN-RAG shows variable performance, sometimes underperforming even compared to the no-retriever baseline, such as on STaRK-Prime using Biomixtral. The "w/o retriever" baseline, representing an LLM without retrieval augmentation, performs surprisingly well in some settings, indicating that strong LLMs alone can capture a significant amount of relevant knowledge. For example, on RiTeK-ADint with Biomixtral, it achieves a Rouge-1 F1 score of 54.15, close to the GNN-RAG. However, in most cases, retrieval-augmented methods still yield superior performance. Notably, Biomixtral 7b tends to outperform the other two LLMs when combined with retrieval, especially in recall and F1. These results suggest that both the choice of retrieval strategy and the backbone LLM significantly impact end-task performance.

\subsubsection{Case Study of Path and Answer Quality}

\begin{figure}[t]
        \centering
        \includegraphics[width=0.8\columnwidth]{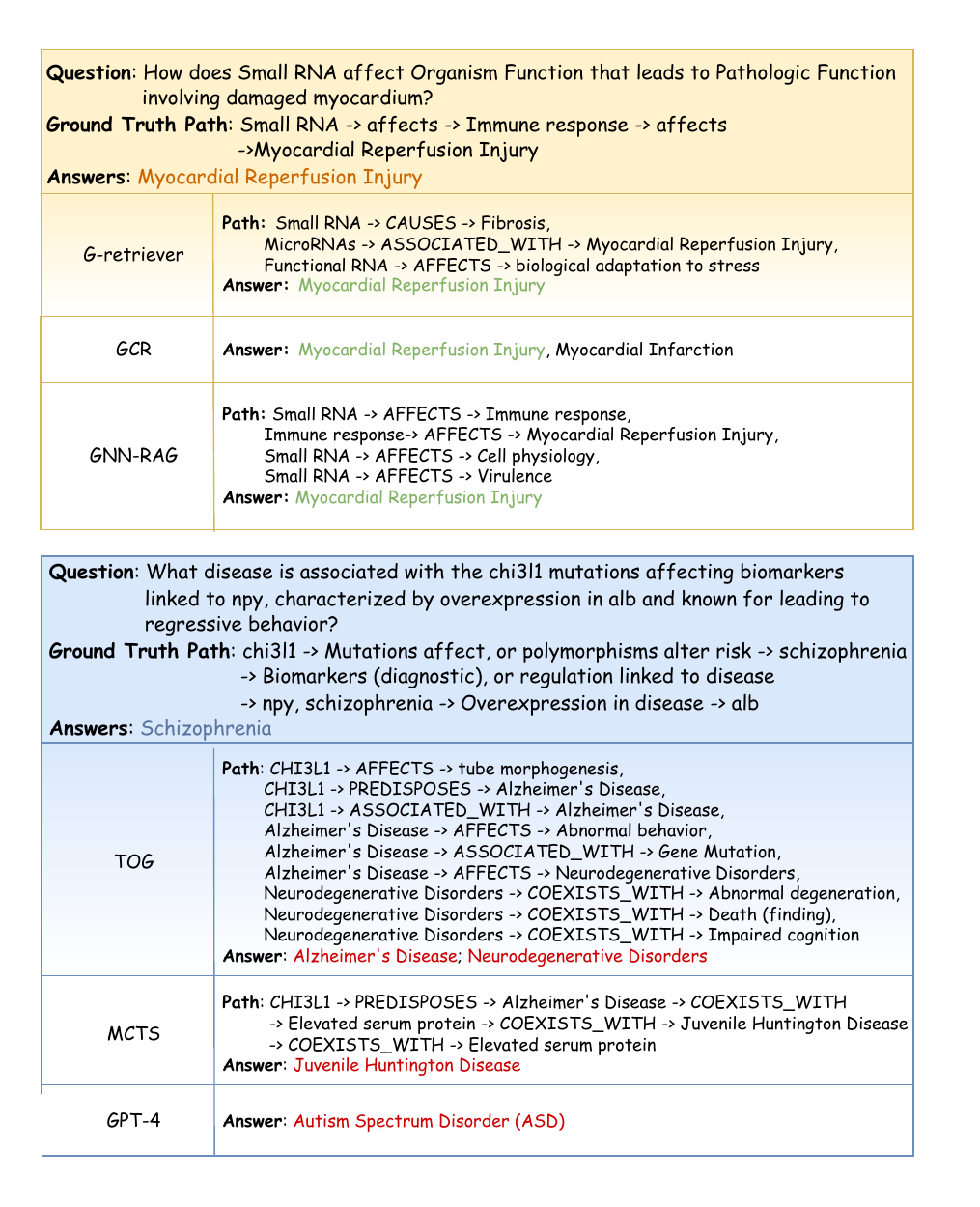}
	\caption{A case study on RiTeK}
	\label{con:case_study_final}
\end{figure}

We conduct a qualitative analysis to compare the reasoning paths and predicted answers from different retrieval models on two biomedical question-answering examples.  As shown in Figure~\ref{con:case_study_final}, all models successfully predicted the correct answer \textit{Myocardial Reperfusion Injury} in the first case, although their reasoning paths varied in granularity and relevance. \textsc{G-retriever} and \textsc{GNN-RAG} produced informative multi-hop paths that partially overlapped with the ground truth.

In contrast, for the second question involving \textit{CHI3L1} and schizophrenia, only the ground truth path led to the correct answer. All baseline models failed: \textsc{TOG} and \textsc{MCTS} generated incorrect reasoning chains centered around \textit{Alzheimer's Disease} and \textit{Juvenile Huntington Disease}, while \textsc{GPT-4} hallucinated \textit{Autism Spectrum Disorder}. These errors reveal the challenge of modeling rare or indirect biomedical associations, especially when entity relations involve subtle phenotypic markers. This case highlights the importance of precise multi-hop reasoning and clinically aligned retrieval in semi-structured biomedical graphs.

\section{Conclusion}



We present RiTeK, the first dataset specifically designed to evaluate the capability of models in handling complex reasoning over textual knowledge graphs (TKGs). This dataset offers diverse topological structures, relational types, entity types, and queries that integrate relational and textual information, requiring sophisticated reasoning across TKGs. RiTeK also includes rich textual descriptions for each node. To ensure the authenticity and accuracy of the queries, medical experts performed stringent validation. RiTeK sets a new standard for evaluating real-world retrieval systems. We evaluated 11 retrieval models on our benchmark dataset. Our experiments on RiTeK reveal significant challenges faced by current models in effectively handling both textual and relational information, especially under complex topological structures involving intricate relations and entities. RiTeK paves the way for future research aimed at advancing retrieval systems by emphasizing the need to enhance reasoning capabilities, particularly in retrieving complex reasoning paths under answer attribute constraints.


\section{Limitations}



RiTeK is currently limited to queries that involve only a single topic entity and rely solely on the textual and structural information present in the graph. Future work should explore the inclusion of multiple topic entities and incorporate additional modalities, such as images, to enable a more comprehensive and robust information retrieval system.


Although we employed four medical
experts for human evaluation, increasing the number of qualified domain experts would improve the
statistical significance and robustness of our findings. Future work should consider expanding the
pool of experts and addressing issues of fairness,
and potential biases inherent in
LLMs.




\bibliography{acl_reference}

@article{wu2024stark,
  title={STaRK: Benchmarking LLM Retrieval on Textual and Relational Knowledge Bases},
  author={Wu, Shirley and Zhao, Shiyu and Yasunaga, Michihiro and Huang, Kexin and Cao, Kaidi and Huang, Qian and Ioannidis, Vassilis N and Subbian, Karthik and Zou, James and Leskovec, Jure},
  journal={arXiv preprint arXiv:2404.13207},
  year={2024}
}

@article{wu2024avatar,
  title={AvaTaR: Optimizing LLM Agents for Tool-Assisted Knowledge Retrieval},
  author={Wu, Shirley and Zhao, Shiyu and Huang, Qian and Huang, Kexin and Yasunaga, Michihiro and Cao, Kaidi and Ioannidis, Vassilis N and Subbian, Karthik and Leskovec, Jure and Zou, James},
  journal={arXiv preprint arXiv:2406.11200},
  year={2024}
}

@incollection{lovasz1993random,
  title={Random walks on graphs: A survey},
  author={Lov{\'a}sz, L{\'a}szl{\'o}},
  booktitle={Combinatorics, Paul Erdős is Eighty},
  volume={2},
  pages={1--46},
  year={1993},
  publisher={János Bolyai Mathematical Society}
}

@article{he2024g,
  title={G-retriever: Retrieval-augmented generation for textual graph understanding and question answering},
  author={He, Xiaoxin and Tian, Yijun and Sun, Yifei and Chawla, Nitesh V and Laurent, Thomas and LeCun, Yann and Bresson, Xavier and Hooi, Bryan},
  journal={arXiv preprint arXiv:2402.07630},
  year={2024}
}

@article{xia2024knowledge,
  title={Knowledge-Aware Query Expansion with Large Language Models for Textual and Relational Retrieval},
  author={Xia, Yu and Wu, Junda and Kim, Sungchul and Yu, Tong and Rossi, Ryan A and Wang, Haoliang and McAuley, Julian},
  journal={arXiv preprint arXiv:2410.13765},
  year={2024}
}

@article{luo2024graph,
  title={Graph-constrained Reasoning: Faithful Reasoning on Knowledge Graphs with Large Language Models},
  author={Luo, Linhao and Zhao, Zicheng and Gong, Chen and Haffari, Gholamreza and Pan, Shirui},
  journal={arXiv preprint arXiv:2410.13080},
  year={2024}
}

@article{yao2024tree,
  title={Tree of thoughts: Deliberate problem solving with large language models},
  author={Yao, Shunyu and Yu, Dian and Zhao, Jeffrey and Shafran, Izhak and Griffiths, Tom and Cao, Yuan and Narasimhan, Karthik},
  journal={Advances in Neural Information Processing Systems},
  volume={36},
  year={2024}
}

@article{mavromatis2024gnn,
  title={GNN-RAG: Graph Neural Retrieval for Large Language Model Reasoning},
  author={Mavromatis, Costas and Karypis, George},
  journal={arXiv preprint arXiv:2405.20139},
  year={2024}
}

@article{sun2023think,
  title={Think-on-graph: Deep and responsible reasoning of large language model with knowledge graph},
  author={Sun, Jiashuo and Xu, Chengjin and Tang, Lumingyuan and Wang, Saizhuo and Lin, Chen and Gong, Yeyun and Shum, Heung-Yeung and Guo, Jian},
  journal={arXiv preprint arXiv:2307.07697},
  year={2023}
}

@inproceedings{Zhang2017Variational,
  title={Variational reasoning for question answering with knowledge graph},
  author={Zhang, Yuyu and Dai, Hanjun and Kozareva, Zornitsa and Smola, Alexander J and Song, Le},
  booktitle={Thirty-Second AAAI Conference on Artificial Intelligence},
  year={2018}
}

@inproceedings{Yih2016The,
  title={The Value of Semantic Parse Labeling for Knowledge Base Question Answering},
  author={Yih, Wen Tau and Richardson, Matthew and Meek, Chris and Chang, Ming Wei and Suh, Jina},
  booktitle={Proceedings of the 54th Annual Meeting of the Association for Computational Linguistics (Volume 2: Short Papers)},
  year={2016}
}

@article{rajpurkar2016squad,
  title={Squad: 100,000+ questions for machine comprehension of text},
  author={Rajpurkar, P},
  journal={arXiv preprint arXiv:1606.05250},
  year={2016}
}

@article{dunn2017searchqa,
  title={Searchqa: A new q\&a dataset augmented with context from a search engine},
  author={Dunn, Matthew and Sagun, Levent and Higgins, Mike and Guney, V Ugur and Cirik, Volkan and Cho, Kyunghyun},
  journal={arXiv preprint arXiv:1704.05179},
  year={2017}
}

@article{joshi2017triviaqa,
  title={Triviaqa: A large scale distantly supervised challenge dataset for reading comprehension},
  author={Joshi, Mandar and Choi, Eunsol and Weld, Daniel S and Zettlemoyer, Luke},
  journal={arXiv preprint arXiv:1705.03551},
  year={2017}
}

@article{trischler2016newsqa,
  title={Newsqa: A machine comprehension dataset},
  author={Trischler, Adam and Wang, Tong and Yuan, Xingdi and Harris, Justin and Sordoni, Alessandro and Bachman, Philip and Suleman, Kaheer},
  journal={arXiv preprint arXiv:1611.09830},
  year={2016}
}

@article{welbl2018constructing,
  title={Constructing datasets for multi-hop reading comprehension across documents},
  author={Welbl, Johannes and Stenetorp, Pontus and Riedel, Sebastian},
  journal={Transactions of the Association for Computational Linguistics},
  volume={6},
  pages={287--302},
  year={2018},
  publisher={MIT Press One Rogers Street, Cambridge, MA 02142-1209, USA journals-info~…}
}

@article{yang2018hotpotqa,
  title={HotpotQA: A dataset for diverse, explainable multi-hop question answering},
  author={Yang, Zhilin and Qi, Peng and Zhang, Saizheng and Bengio, Yoshua and Cohen, William W and Salakhutdinov, Ruslan and Manning, Christopher D},
  journal={arXiv preprint arXiv:1809.09600},
  year={2018}
}

@article{jin2021disease,
  title={What disease does this patient have? a large-scale open domain question answering dataset from medical exams},
  author={Jin, Di and Pan, Eileen and Oufattole, Nassim and Weng, Wei-Hung and Fang, Hanyi and Szolovits, Peter},
  journal={Applied Sciences},
  volume={11},
  number={14},
  pages={6421},
  year={2021},
  publisher={MDPI}
}

@article{jin2019pubmedqa,
  title={Pubmedqa: A dataset for biomedical research question answering},
  author={Jin, Qiao and Dhingra, Bhuwan and Liu, Zhengping and Cohen, William W and Lu, Xinghua},
  journal={arXiv preprint arXiv:1909.06146},
  year={2019}
}

@article{hendrycks2020measuring,
  title={Measuring massive multitask language understanding},
  author={Hendrycks, Dan and Burns, Collin and Basart, Steven and Zou, Andy and Mazeika, Mantas and Song, Dawn and Steinhardt, Jacob},
  journal={arXiv preprint arXiv:2009.03300},
  year={2020}
}

@article{yao2024medqa,
  title={MedQA-CS: Benchmarking Large Language Models Clinical Skills Using an AI-SCE Framework},
  author={Yao, Zonghai and Zhang, Zihao and Tang, Chaolong and Bian, Xingyu and Zhao, Youxia and Yang, Zhichao and Wang, Junda and Zhou, Huixue and Jang, Won Seok and Ouyang, Feiyun and others},
  journal={arXiv preprint arXiv:2410.01553},
  year={2024}
}

@article{zheng2021pharmkg,
  title={PharmKG: a dedicated knowledge graph benchmark for bomedical data mining},
  author={Zheng, Shuangjia and Rao, Jiahua and Song, Ying and Zhang, Jixian and Xiao, Xianglu and Fang, Evandro Fei and Yang, Yuedong and Niu, Zhangming},
  journal={Briefings in bioinformatics},
  volume={22},
  number={4},
  pages={bbaa344},
  year={2021},
  publisher={Oxford University Press}
}

@article{xiao2024repurposing,
  title={Repurposing non-pharmacological interventions for Alzheimer's disease through link prediction on biomedical literature},
  author={Xiao, Yongkang and Hou, Yu and Zhou, Huixue and Diallo, Gayo and Fiszman, Marcelo and Wolfson, Julian and Zhou, Li and Kilicoglu, Halil and Chen, You and Su, Chang and others},
  journal={Scientific reports},
  volume={14},
  number={1},
  pages={8693},
  year={2024},
  publisher={Nature Publishing Group UK London}
}

@article{li2022semantic,
  title={Semantic structure based query graph prediction for question answering over knowledge graph},
  author={Li, Mingchen and Ji, Shihao},
  journal={arXiv preprint arXiv:2204.10194},
  year={2022}
}

@article{wei2022chain,
  title={Chain-of-thought prompting elicits reasoning in large language models},
  author={Wei, Jason and Wang, Xuezhi and Schuurmans, Dale and Bosma, Maarten and Xia, Fei and Chi, Ed and Le, Quoc V and Zhou, Denny and others},
  journal={Advances in neural information processing systems},
  volume={35},
  pages={24824--24837},
  year={2022}
}

@article{achiam2023gpt,
  title={Gpt-4 technical report},
  author={Achiam, Josh and Adler, Steven and Agarwal, Sandhini and Ahmad, Lama and Akkaya, Ilge and Aleman, Florencia Leoni and Almeida, Diogo and Altenschmidt, Janko and Altman, Sam and Anadkat, Shyamal and others},
  journal={arXiv preprint arXiv:2303.08774},
  year={2023}
}

@article{cohan2016revisiting,
  title={Revisiting summarization evaluation for scientific articles},
  author={Cohan, Arman and Goharian, Nazli},
  journal={arXiv preprint arXiv:1604.00400},
  year={2016}
}

@inproceedings{li2023understand,
  title={Understand the dynamic world: An end-to-end knowledge informed framework for open domain entity state tracking},
  author={Li, Mingchen and Huang, Lifu},
  booktitle={Proceedings of the 46th International ACM SIGIR Conference on Research and Development in Information Retrieval},
  pages={842--851},
  year={2023}
}

@article{chaslot2010monte,
  title={Monte-carlo tree search},
  author={Chaslot, Guillaume Maurice Jean-Bernard Chaslot},
  year={2010}
}

@inproceedings{gu2021beyond,
  title={Beyond IID: three levels of generalization for question answering on knowledge bases},
  author={Gu, Yu and Kase, Sue and Vanni, Michelle and Sadler, Brian and Liang, Percy and Yan, Xifeng and Su, Yu},
  booktitle={Proceedings of the Web Conference 2021},
  pages={3477--3488},
  year={2021}
}

@inproceedings{bao2016constraint,
  title={Constraint-based question answering with knowledge graph},
  author={Bao, Junwei and Duan, Nan and Yan, Zhao and Zhou, Ming and Zhao, Tiejun},
  booktitle={Proceedings of COLING 2016, the 26th international conference on computational linguistics: technical papers},
  pages={2503--2514},
  year={2016}
}

@inproceedings{trivedi2017lc,
  title={Lc-quad: A corpus for complex question answering over knowledge graphs},
  author={Trivedi, Priyansh and Maheshwari, Gaurav and Dubey, Mohnish and Lehmann, Jens},
  booktitle={The Semantic Web--ISWC 2017: 16th International Semantic Web Conference, Vienna, Austria, October 21-25, 2017, Proceedings, Part II 16},
  pages={210--218},
  year={2017},
  organization={Springer}
}

@inproceedings{besta2024graph,
  title={Graph of thoughts: Solving elaborate problems with large language models},
  author={Besta, Maciej and Blach, Nils and Kubicek, Ales and Gerstenberger, Robert and Podstawski, Michal and Gianinazzi, Lukas and Gajda, Joanna and Lehmann, Tomasz and Niewiadomski, Hubert and Nyczyk, Piotr and others},
  booktitle={Proceedings of the AAAI Conference on Artificial Intelligence},
  volume={38},
  number={16},
  pages={17682--17690},
  year={2024}
}

@misc{lin2024biokgbench,
      title={BioKGBench: A Knowledge Graph Checking Benchmark of AI Agent for Biomedical Science}, 
      author={Xinna Lin and Siqi Ma and Junjie Shan and Xiaojing Zhang and Shell Xu Hu and Tiannan Guo and Stan Z. Li and Kaicheng Yu},
      year={2024},
      eprint={2407.00466},
      archivePrefix={arXiv},
      primaryClass={cs.CL},
      url={https://arxiv.org/abs/2407.00466}, 
}

@incollection{yan2024bridging,
title={Bridging the Gap: Generating a Comprehensive Biomedical Knowledge Graph Question Answering Dataset},
author={Yan, Xi and Westphal, Patrick and Seliger, Jan and Usbeck, Ricardo},
booktitle={ECAI 2024},
pages={1198--1205},
year={2024},
publisher={IOS Press}
}

\appendix

\section{Appendix}

\subsection{Ethics Statement}
All experiments in this study were conducted using publicly available datasets, including Prime, ADint, and PharmKG. These datasets contain only de-identified, non-personal, and non-sensitive information that was released for research purposes under appropriate data licenses. No private or confidential patient data were accessed or used. Consequently, no additional ethical approval was required.  We employed AI tools to assist with grammar revision.
\subsection{Hyperparameters}
In the retrieval model, we set the maximum search depth to $d=3$ for both Random Walk and MCTS. The number of rollouts in MCTS is 16. For G-Retriever, KAR, GNN-RAG, and GCR, we adopted the same hyperparameters as reported in their respective source papers.

\subsection{TKG resources}
Ensembl~\footnote{https://useast.ensembl.org/index.html}, UMLS~\footnote{https://www.nlm.nih.gov/research/umls/index.html}, and Mondo Disease Ontology~\footnote{https://mondo.monarchinitiative.org/}.

\subsection{Medical textual knowledge graph construction}
\label{Medical_TKGs}
We construct two medical TKGs based on PharmKG~\cite{zheng2021pharmkg} and ADint~\cite{xiao2024repurposing}, as the increased number of entity and relation types introduces significant challenges for path retrieval in the question answering over textual knowledge graph.
We present the statistics of the relational
structure in Table \ref{con:data_statisitics_meidcal_sskg} and introduce each TKG as follows:

\textbf{PharmKG Textual Knowledge Graph}: We leverage the existing  medical knowledge  graph  PharmKG~\cite{zheng2021pharmkg} which is a multi-relational, attribute-rich biomedical knowledge graph (KG) constructed from six publicly available databases that provide high-quality structured information. These databases include OMIM, DrugBank, PharmGKB, Therapeutic Target Database (TTD), SIDER, and HumanNet. PharmKG consists of over 500,000 distinct interconnections between genes, drugs, and diseases, encompassing 29 types of relationships within a vocabulary of approximately 8,000 disambiguated entities.  To enhance the entity attributes, we incorporate textual details from various databases, including Ensembl, UMLS, and Mondo Disease Ontology, as supplementary data sources.

\textbf{ADInt Textual Knowledge Graph}: ADInt\cite{xiao2024repurposing} is a comprehensive knowledge graph (KG) constructed from biomedical literature, focusing on non-pharmacological interventions (NPI) and their associations with Alzheimer’s disease (AD). ADInt includes 162,212 entities spanning 113 UMLS semantic types, which, upon further classification, consist of 25,604 drugs, 16,474 diseases, 46,060 genes and proteins, 2,525 dietary supplements (DS), and 128 complementary and integrative health (CIH) interventions. Moreover, ADInt contains 1,017,284 triples, capturing 15 distinct relation types, offering a rich dataset for exploring the intricate relationships between NPIs and AD.    Same as PharmKG, we  also incorporate textual details from various databases, including Ensembl, UMLS, and Mondo Disease Ontology, as supplementary data sources.

\subsection{The prompt of Combining Textual and Relational Information}
\label{con:prompt of Combining Textual and Relational Information}
You are a creative assistant tasked with generating natural, diverse, and realistic queries by combining textual properties and relational templates. 
Write the query from the perspective of a <persona>, ensuring it is concise, human-like, and paraphrased while retaining the original meaning. 

Consider the following characteristics for the persona:
\begin{itemize}
\item Doctor: Formulate direct and practical questions aimed at diagnosing and treating. These questions should focus on side effects, symptoms, complications, and other clinically relevant aspects.
\item Medical Scientist: Generate detailed and specific questions reflecting the complexity of scientific inquiry. These questions should explore etiology, pathophysiology, genetic factors, pathways, proteins, or molecular functions.
\item Patient: Create straightforward questions that avoid professional medical terminology. These questions should focus on practical concerns, such as symptoms, effects, inheritance, or other relatable aspects, and may include more context from daily life.
\end{itemize}

\textbf{Textual Properties}: [<input\_textual\_properties>]

\textbf{Relational Templates}: [<input\_relational\_templates>]

\textbf{Persona}: <input\_persona> (e.g., Doctor, Medical Scientist, Patient)

Ensure the query is realistic and diverse, leveraging flexibility in how the textual and relational elements are presented. Avoid directly copying the input phrases; instead, paraphrase them while retaining their original meaning.
Please output only the generated query without any additional comments or explanations.





\subsection{Data Analysis of query length and answer length}
\label{Data_Analysis_of_QA_length}

\begin{figure}
    \vspace{-5mm}
    \centering
    \includegraphics[width=0.46\textwidth]{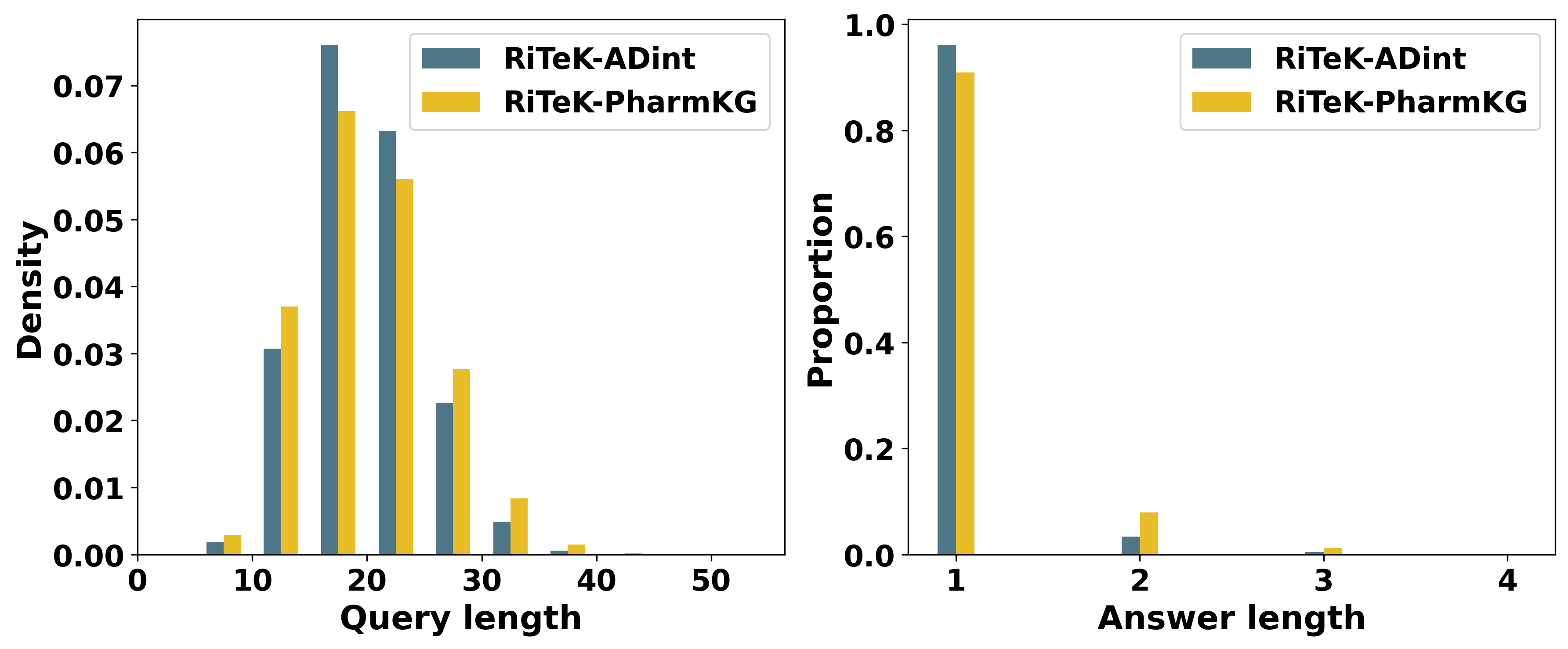}
     \vspace{-3mm}
    \caption{Distribution of query lengths and answer lengths on RiTeK-ADint and RiTeK-PharmKG datasets}
     \vspace{-5mm}
    \label{fig:sample_image}
\end{figure}

We analyzed the distribution of query lengths (i.e., the number of words in each query) to assess the complexity of the queries and the amount of information they contain. As shown in the Figure \ref{fig:sample_image}, the query lengths range from 5 to 40 words, with approximately 69\% and 61\% of queries in the two datasets having lengths between 15 and 25 words.

Then, we analyzed the proportion of ground truth answers associated with each query. Generally, the more ground truth answers there are, the less precise the textual requirements in the query tend to be. To increase the difficulty of the question-answering task, we filtered out queries with too many ground truth answers during the dataset creation process, retaining only those with a maximum of three ground truth answers. In both datasets, over 90\% of queries have a single ground truth answer, indicating that our queries are enriched with detailed textual information from entity attributes. This introduces more challenges when developing new graph retrieval methods

\section{Relational Template }
\subsection{RiTeK-PharmKG}
\label{relation template in RiTeK-PharmKG}

\begin{enumerate}
     \item  Gene -> [Production by cell population] -> Gene
     \item Gene -> [Enhance response, or activate, stimulate] -> Gene
  \item   Gene -> [Relationships involving regulation and pathways] -> Gene
 \item  Gene -> [Binding, ligand] -> Gene
 \item  Gene -> [Affects expression/production] -> Gene
\item Gene -> [Gene-Gene] -> Gene
 \item Chemical -> [Chemical-Chemical] -> Chemical
\item Disease -> [Ancestors of disease] -> Disease
\item Disease -> [Associations between diseases] -> Disease
\item Gene -> [Interactions] -> Chemical
\item Chemical -> [Interactions] -> Gene
\item Gene -> [Interactions] -> Gene
\item Gene -> [Interactions] -> Disease
\item Gene -> [Drug targets] -> Disease
\item Gene -> [Role in pathogenesis, or promotes progression] -> Disease
\item Gene -> [Mutations affect, or polymorphisms alter risk] -> Disease
\item Disease -> [Biomarkers (diagnostic), or regulation linked to disease] -> Gene
\item Disease -> [Overexpression in disease] -> Gene
\item Chemical -> [Treatment or therapy] -> Disease
\item Chemical -> [Side effect or adverse event] -> Disease
\item Chemical -> [Inhibits cell growth] -> Disease
\item Chemical -> [Role in pathogenesis] -> Disease
\item Chemical -> [Prevents, suppresses, or alleviates, reduces] -> Disease
\item Disease -> [Biomarkers (progression)] -> Chemical
\item Chemical -> [Agonism, activation, or antagonism, blocking] -> Gene
\item Chemical -> [Binding, ligand] -> Gene
\item Chemical -> [Affects expression/production] -> Gene
\item Chemical -> [Inhibits] -> Gene
\item Gene -> [Transport, channels] -> Chemical
\item Gene -> [Metabolism, pharmacokinetics] -> Chemical
\item Gene -> [Enzyme activity] -> Chemical
\item Gene -> [Enhance response, or activate, stimulate] -> Gene -> [Drug targets] -> Disease
\item Gene -> [Enhance response, or activate, stimulate] -> Gene -> [Role in pathogenesis, or promotes progression] -> Disease
\item Gene -> [Enhance response, or activate, stimulate] -> Gene -> [Mutations affect, or polymorphisms alter risk] -> Disease
\item Gene -> [Relationships involving regulation and pathways] -> Gene -> [Binding, ligand] -> Gene
\item Gene -> [Binding, ligand] -> Gene -> [Affects expression/production] -> Gene
\item Gene -> [Interactions] -> Gene -> [Interactions] -> Chemical
\item Gene -> [Transport, channels] -> Chemical -> [Agonism, activation, or antagonism, blocking] -> Gene
\item Gene -> [Metabolism, pharmacokinetics] -> Chemical -> [Binding, ligand] -> Gene
\item Gene -> [Enhance response, or activate, stimulate] -> Gene -> [Enhance response, or activate, stimulate] -> Gene
\item Gene -> [Interactions] -> Chemical -> [Treatment or therapy] -> Disease
\item Gene -> [Interactions] -> Chemical -> [Side effect or adverse event] -> Disease
\item Gene -> [Interactions] -> Disease -> [Biomarkers (diagnostic), or regulation linked to disease] -> Gene
\item Chemical -> [Treatment or therapy] -> Disease -> [Biomarkers (diagnostic), or regulation linked to disease] -> Gene
\item Disease -> [Associations between diseases] -> Disease -> [Ancestors of disease] -> Disease
\item Disease -> [Biomarkers (diagnostic), or regulation linked to disease] -> Disease -> [Biomarkers (diagnostic), or regulation linked to disease] -> Gene
\item Gene -> [Interactions] -> Gene -> [Transport, channels] -> Chemical
\item Gene -> [Metabolism, pharmacokinetics] -> Chemical -> [Binding, ligand] -> Gene
\item Gene -> [Enhance response, or activate, stimulate] -> Gene -> [Drug targets] -> Disease -> [Biomarkers (diagnostic), or regulation linked to disease] -> Gene
\item Gene -> [Enhance response, or activate, stimulate] -> Gene -> [Mutations affect, or polymorphisms alter risk] -> Disease -> [Overexpression in disease] -> Gene
\item Gene -> [Transport, channels] -> Chemical -> [Agonism, activation, or antagonism, blocking] -> Gene -> [Binding, ligand] -> Chemical
\item Gene -> [Metabolism, pharmacokinetics] -> Chemical -> [Binding, ligand] -> Gene -> [Inhibits] -> Chemical
\item Gene -> [Interactions] -> Chemical -> [Treatment or therapy] -> Disease -> [Biomarkers (diagnostic), or regulation linked to disease] -> Gene
\item Gene -> [Interactions] -> Disease -> [Biomarkers (diagnostic), or regulation linked to disease] -> Gene -> [Transport, channels] -> Chemical
\item Gene -> [Role in pathogenesis, or promotes progression] -> Disease -> [Biomarkers (diagnostic), or regulation linked to disease] -> Gene -> [Metabolism, pharmacokinetics] -> Chemical
\item Chemical -> [Agonism, activation, or antagonism, blocking] -> Gene -> [Drug targets] -> Disease -> [Biomarkers (diagnostic), or regulation linked to disease] -> Gene
\item Disease -> [Biomarkers (diagnostic), or regulation linked to disease] -> Disease -> [Biomarkers (diagnostic), or regulation linked to disease] -> Gene -> [Role in pathogenesis, or promotes progression] -> Disease
\item Disease -> [Biomarkers (diagnostic), or regulation linked to disease] -> Gene -> [Metabolism, pharmacokinetics] -> Chemical -> [Side effect or adverse event] -> Disease
\item Gene -> [Production by cell population] -> Gene -> [Enhance response, or activate, stimulate] -> Gene -> [Relationships involving regulation and pathways] -> Gene
\item Gene -> [Enhance response, or activate, stimulate] -> Gene -> [Binding, ligand] -> Gene -> [Affects expression/production] -> Gene
\item Gene -> [Relationships involving regulation and pathways] -> Gene -> [Gene-Gene] -> Gene -> [Binding, ligand] -> Gene
\item Gene -> [Interactions] -> Gene -> [Interactions] -> Gene -> [Transport, channels] -> Chemical
\item Gene -> [Interactions] -> Gene -> [Interactions] -> Gene -> [Metabolism, pharmacokinetics] -> Chemical
\item Gene -> [Enhance response, or activate, stimulate] -> Gene -> [Mutations affect, or polymorphisms alter risk] -> Disease -> [Overexpression in disease] -> Gene
\item Gene -> [Enzyme activity] -> Chemical -> [Affects expression/production] -> Gene -> [Chemical-Chemical] -> Chemical
\item Gene -> [Interactions] -> Chemical -> [Role in pathogenesis] -> Disease -> [Overexpression in disease] -> Gene
\item Chemical -> [Side effect or adverse event] -> Disease -> [Biomarkers (diagnostic), or regulation linked to disease] -> Gene -> [Mutations affect, or polymorphisms alter risk] -> Disease
\item Chemical -> [Inhibits cell growth] -> Disease -> [Overexpression in disease] -> Gene -> [Role in pathogenesis, or promotes progression] -> Disease

\end{enumerate}

\subsection{RiTeK-ADint}

\label{relation_tempalte_RiTeK-ADint}

\begin{enumerate}
    \item Amino Acid, Peptide, or Protein -> [affects] -> Cell Function
\item Amino Acid, Peptide, or Protein -> [affects] -> Disease or Syndrome
\item Amino Acid, Peptide, or Protein -> [causes] -> Anatomical Abnormality
\item Amino Acid, Peptide, or Protein -> [interacts with] -> Pharmacologic Substance
\item Anatomical Abnormality -> [affects] -> Organ or Tissue Function
\item Anatomical Abnormality -> [complicates] -> Disease or Syndrome
\item Anatomical Abnormality -> [manifestation of] -> Genetic Function
\item Antibiotic -> [affects] -> Molecular Function
\item Antibiotic -> [causes] -> Pathologic Function
\item Antibiotic -> [disrupts] -> Cell Component
\item Antibiotic -> [treats] -> Disease or Syndrome
\item Bacterium -> [causes] -> Cell or Molecular Dysfunction
\item Bacterium -> [interacts with] -> Human
\item Biologically Active Substance -> [affects] -> Organism Function
\item Biologically Active Substance -> [causes] -> Injury or Poisoning
\item Biologically Active Substance -> [disrupts] -> Gene or Genome
\item Body Part, Organ, or Organ Component -> [produces] -> Immunologic Factor
\item Cell Component -> [affects] -> Molecular Function
\item Cell Component -> [produces] -> Nucleic Acid, Nucleoside, or Nucleotide
\item Cell Function -> [affects] -> Mental or Behavioral Dysfunction
\item Cell Function -> [produces] -> Biologically Active Substance
\item Cell or Molecular Dysfunction -> [affects] -> Neoplastic Process
\item Cell or Molecular Dysfunction -> [manifestation of] -> Pathologic Function
\item Cell -> [produces] -> Organic Chemical
\item Congenital Abnormality -> [affects] -> Virus
\item Congenital Abnormality -> [manifestation of] -> Organism Function
\item Diagnostic Procedure -> [affects] -> Genetic Function
\item Disease or Syndrome -> [affects] -> Organ or Tissue Function
\item Disease or Syndrome -> [associated with] -> Therapeutic or Preventive Procedure
\item Disease or Syndrome -> [manifestation of] -> Cell or Molecular Dysfunction
\item Finding -> [manifestation of] -> Pathologic Function
\item Gene or Genome -> [produces] -> Amino Acid, Peptide, or Protein
\item Genetic Function -> [affects] -> Human
\item Genetic Function -> [produces] -> Cell Component
\item Hazardous or Poisonous Substance -> [affects] -> Mental or Behavioral Dysfunction
\item Hazardous or Poisonous Substance -> [disrupts] -> Organ or Tissue Function
\item Health Care Activity -> [affects] -> Disease or Syndrome
\item Human -> [interacts with] -> Human
\item Immunologic Factor -> [affects] -> Pathologic Function
\item Indicator, Reagent, or Diagnostic Aid -> [interacts with] -> Hazardous or Poisonous Substance
\item Injury or Poisoning -> [disrupts] -> Genetic Function
\item Medical Device -> [treats] -> Mental or Behavioral Dysfunction
\item Mental or Behavioral Dysfunction -> [affects] -> Organism Function
\item Molecular Function -> [affects] -> Virus
\item Neoplastic Process -> [affects] -> Bacterium
\item Neoplastic Process -> [associated with] -> Neoplastic Process
\item Nucleic Acid, Nucleoside, or Nucleotide -> [interacts with] -> Immunologic Factor
\item Organ or Tissue Function -> [produces] -> Immunologic Factor
\item Organic Chemical -> [affects] -> Pathologic Function
\item Organic Chemical -> [interacts with] -> Pharmacologic Substance
\item Organism Function -> [affects] -> Disease or Syndrome
\item Pathologic Function -> [associated with] -> Therapeutic or Preventive Procedure
\item Pathologic Function -> [manifestation of] -> Organ or Tissue Function
\item Pharmacologic Substance -> [affects] -> Genetic Function
\item Pharmacologic Substance -> [treats] -> Sign or Symptom
\item Sign or Symptom -> [manifestation of] -> Genetic Function
\item Therapeutic or Preventive Procedure -> [affects] -> Neoplastic Process
\item Virus -> [interacts with] -> Human
\end{enumerate}

\section{Topological Structures}
\label{con:Topological_Structures}
\subsection{Definition of Six Reasoning Topologies}
Following the semantic structure framework introduced by \cite{li2022semantic}, we define six reasoning topologies that serve as the structural backbone for question generation and reasoning path simulation in RiTeK.
Each topology represents a distinct reasoning pattern between entities in a textual knowledge graph (TKG), reflecting different levels of relational complexity and logical dependency.
\begin{itemize}
\item 1-hop: The simplest reasoning structure, consisting of a single relation connecting the topic entity and the answer entity.
\item 2-hop: A two-step linear reasoning chain where the answer is connected to the topic entity through an intermediate entity.
\item 3-hop: A longer reasoning chain with three relational steps, representing more complex dependency and multi-level inference.
\item 1-hop with constraint: A single relational edge combined with a semantic or categorical constraint that filters valid answers.
\item 2-hop with constraint: A two-hop reasoning chain where the final answer is subject to an additional semantic or categorical restriction.
\item Two-to-one (Converging paths): Two distinct entities or relations converge on a common target entity. This topology reflects intersective reasoning, where the answer satisfies multiple relational constraints simultaneously.
\end{itemize}
\subsection{Examples in the Medical Domain}
\begin{itemize}
\item 1-hop: For example, Amino Acid, Peptide, or Protein → [affects] → Disease or Syndrome illustrates a direct relationship such as “Which diseases are affected by a given protein?”

\item 2-hop: An example is Gene → [Interactions] → Chemical → [Treatment or therapy] → Disease, corresponding to the question “Which diseases can be treated by chemicals that interact with a specific gene?”

\item 3-hop: For instance, Gene → [Interactions] → Chemical → [Treatment or therapy] → Disease → [Biomarkers (diagnostic), or regulation linked to disease] → Gene models complex interactions such as “Which genes serve as biomarkers for diseases treated by chemicals interacting with a given gene?”

\item 1-hop with constraint: For example, Pharmacologic Substance → [treats] → Disease or Syndrome under the constraint “Disease type = neurodegenerative” represents questions like “Which pharmacologic substances treat neurodegenerative diseases?”

\item 2-hop with constraint: An example is Gene → [Enhance response, or activate, stimulate] → Gene → [Drug targets] → Disease with the constraint “Disease subtype = inflammatory”, corresponding to “Which genes activate other genes that target inflammatory diseases?”

\item Two-to-one (Converging paths): For instance, Gene 1 → [Interactions] → Chemical ← [Interactions] ← Gene 2 captures the question “Which chemicals interact with both Gene A and Gene B?”.
\end{itemize}


\section{Data Distribution Analysis}
\label{con:Data_analysis}
\begin{table}
	\centering
  \vspace{-5mm}
	\renewcommand\arraystretch{1.2} 
 \scalebox{0.6}{
	\begin{tabular} {c|cc} 
		\hline
		& Shannon Entropy  & Type-Token Ratio  \\ 
            \hline
            & \multicolumn{2}{c}{Medical domain}\\
             \hline
            RiTeK-ADint &10.04 & 0.187  \\  
  	    RiTeK-PharmKG &9.61& 0.157  \\
            STARK-PRIME & 9.63  & 0.143  \\
		\hline
            & \multicolumn{2}{c}{General domain}\\
             \hline
		STARK-AMAZON &  10.39  & 0.179  \\
  STARK-MAG & 10.25  & 0.180  \\
		
		\hline
	\end{tabular}}
  \vspace{-2mm}
	\caption{Query diversity measurement}
  \vspace{-5mm}
	\label{con:data_statistics_medical_sskg}
\end{table}

\end{document}